\documentclass{article}
\newif\ifanon
\anonfalse
\ifanon
  \usepackage{tmlr}
\else
  \usepackage[preprint]{tmlr}
\fi
\usepackage[utf8]{inputenc}
\usepackage[T1]{fontenc}
\usepackage{amsmath,amssymb,booktabs,array,graphicx,xcolor,hyperref}
\graphicspath{{}}

\newcommand{\Hun}{$0.467$}
\newcommand{\Hdif}{$0.546$}
\newcommand{\Hval}{$0.698$}
\newcommand{\Hpost}{$0.921$}
\newcommand{\AunoDelta}{$+0.151$}
\newcommand{\RoutingExpl}{$+0.79$}
\newcommand{\VGBcero}{$+0.0002\pm0.0004$}
\newcommand{\LLMtecho}{$+0.0236$} \newcommand{\LLMtechoIC}{$[+0.0150,\,+0.0326]$}
\newcommand{\LLMtechoTbaja}{$+0.002$}
\newcommand{\LLMmodalerr}{$0.833$}
\newcommand{\LLMepsilon}{$0.08$}
\newcommand{\NGunoDelta}{$+0.0007$} \newcommand{\NGunoIC}{$[-0.0065,\,+0.0079]$}
\newcommand{\NbLHuno}{$+0.1312$} \newcommand{\NbLHunoIC}{$[+0.1124,\,+0.1502]$}
\newcommand{\NbLHunoSuave}{$+0.019$}
\newcommand{\NbLRangoCliff}{$0.278$} \newcommand{\NbLRangoSuave}{$0.055$}
\newcommand{\NbLContrCliff}{$+0.198$} \newcommand{\NbLContrSuave}{$+0.031$}
\newcommand{\NbLFracCliff}{$0.759$} \newcommand{\NbLFracSuave}{$0.763$}
\newcommand{\NbLRdosCliff}{$0.26$} \newcommand{\NbLRdosSuave}{$0.75$}
\newcommand{\NdosESS}{$0.40$} \newcommand{\NdosBlind}{$+0.147$}
\newcommand{\FtresARtoque}{$-0.011$} \newcommand{\FtresARIC}{$[-0.068,\,+0.045]$}
\newcommand{\FdosDz}{$4.06$}
\newcommand{\VtresCorr}{$0.63$} \newcommand{\VtresCorrBase}{$0.20$}
\newcommand{\Ytotal}{$+0.430$} \newcommand{\YtotalIC}{$[+0.397,\,+0.463]$}
\newcommand{\Yregimen}{$+0.383$} 
\newcommand{\Yinfo}{$+0.047$} \newcommand{\YinfoIC}{$[+0.029,\,+0.065]$}
\newcommand{\Rlectura}{$+0.383$} \newcommand{\RlecturaIC}{$[+0.341,\,+0.424]$}
\newcommand{\Rregimen}{$+0.000$} \newcommand{\RregimenIC}{$[0.000,\,0.000]$}
\newcommand{\Rasign}{$+0.0011$} \newcommand{\RasignIC}{$[+0.0003,\,+0.0019]$}
\newcommand{\Rforma}{$+0.267$} \newcommand{\RformaIC}{$[+0.246,\,+0.288]$}
\newcommand{\Rdesplaz}{$1.44$}
\newcommand{\FracDif}{$-0.004$} \newcommand{\FracDifIC}{$[-0.150,\,+0.241]$}
\newcommand{\RatioRangos}{$5.1\times$} \newcommand{\RatioRangosIC}{$[3.4,\,8.2]$}
\newcommand{\GsmAccUno}{$0.905$} \newcommand{\GsmPend}{$+0.030$}
\newcommand{\GsmModalErr}{$0.066$} \newcommand{\GsmEps}{$+0.029$}

\title{Where Cognition Lives:\\ Dissecting Emergent from Computed Function\\
in a Minimal Complete Cognitive Architecture}
\author{\name Francisco M. Arrabal-Campos$^{\ast}$ \email fmarrabal@ual.es\\
\addr Department of Chemistry and Physics, Research Centre CIAIMBITAL\\
\addr Department of Engineering, Research Centre CIAIMBITAL\\
\addr University of Almer\'ia, 04120 Almer\'ia, Spain \quad ($^{\ast}$corresponding author)
\AND
\name Francisco G. Montoya \email pagilm@ual.es\\
\addr Department of Engineering, Research Centre CIAIMBITAL\\
\addr University of Almer\'ia, 04120 Almer\'ia, Spain
\AND
\name Alfredo Alcayde \email aalcayde@ual.es\\
\addr Department of Engineering, Research Centre CIAIMBITAL\\
\addr University of Almer\'ia, 04120 Almer\'ia, Spain
\AND
\name Ignacio Fern\'andez \email ifernan@ual.es\\
\addr Department of Chemistry and Physics, Research Centre CIAIMBITAL\\
\addr University of Almer\'ia, 04120 Almer\'ia, Spain}

\ifanon
  \newcommand{\compcite}{\citep{anon2026governor}}
\else
  \newcommand{\compcite}{\citep{arrabal2026governor}}
\fi
\newcommand{\coderepo}{\url{https://github.com/fmarrabal/miuracognitive}}
\begin{document}
\maketitle

\begin{abstract}
A cognitive architecture is more than the module that reasons. It must also
decide how long to think, what deserves the effort, and when looking is better
than planning. We built a minimal but complete system---a recurrent reasoner
with adaptive halting, a homeostatic control field, and a value module---and
spent a research program asking one question of every part: does this function
\emph{emerge} from gradient descent, or must it be \emph{computed} by explicit
machinery? The answers are sharp. Competence emerges. Stopping emerges too,
and appears to be worth more than everything decidable in advance---but that
appearance is instrumentation: in our information hierarchy,
payoff at closely matched mean compute climbs from \Hun{} (uniform
allocation) through \Hdif{} (difficulty) to \Hval{} (ex-ante value), which
matches the class-stake oracle to three decimals---an equality that is partly
by construction, since our stake sensor is perfect. The further climb to \Hpost{}
(posterior self-observation) does \emph{not} survive audit: PonderNet-style
halting returns a halting-weighted \emph{mixture} of hidden states while every
forced-depth baseline returns a single state, and the language head is trained
on the mixture alone. Equalizing the readout annihilates the apparent
advantage of native execution (residual \Rregimen{} \RregimenIC), and with the
readout held fixed and the budget matched, knowing which instance needs how
much compute is worth \Rasign{} \RasignIC. Adding value on top of the
posterior likewise buys nothing (\VGBcero). Value does \emph{not} emerge---trained couplings
capture zero of an available payoff that an explicit allocator captures
completely (\AunoDelta{}, routing correlation \RoutingExpl)---while
anticipation has no payoff to capture at all in these families ($\le+0.001$
across $31$ configurations), so the second-order decisions that do pay must be
computed, at least where value is orthogonal to content as it is here by
construction. On a frozen LLM actuator the same instruments show
that the standard test-time lever, self-consistency voting, is a measured
bound (\LLMtecho{} \LLMtechoIC) and that inter-sample agreement is nearly
worthless as a stopping signal---its mass concentrates on wrong
answers---so watching oneself think is nearly worthless in \emph{both}
regimes, a convergence that was invisible while one of the two was measured
through a readout the other did not share. Almost every negative result in the
program carries its mechanism---the exceptions are declared as
unadjudicated---and the protocol that produced
them---preregistration, adversarial panels, kill-gates, replication rules, and
positive controls for every null we assert---is part of the contribution. We close by
executing our own falsifiable prediction. In a cliff-cost family, where the
compute an instance needs is invisible until it is spent, value under
commitment pays \NbLHuno{} \NbLHunoIC{}, roughly seven times the point
estimate (\NbLHunoSuave{}) in a smooth
family---yet not because the cliff shifts information toward ex-ante decisions
(the ex-ante fraction of the attainable range is \NbLFracCliff{} versus
\NbLFracSuave{}; difference \FracDif{} \FracDifIC{}, a post-hoc
observation compatible with equality but imprecisely estimated), but
because it multiplies the attainable range fivefold (\RatioRangos{}
\RatioRangosIC). On this evidence the magnitude of an allocation problem and
the structure of its information behave as separate axes---the first measured
decisively, the second only bounded.
\end{abstract}

\section{Introduction}\label{sec:intro}

Two intuitions about machine cognition pull in opposite directions. One says
that everything worth having emerges from scale and gradient
descent~\citep{wei2022emergent}---and that architectural commitments mostly
get in the way---though what truly emerges is itself
contested~\citep{schaeffer2023mirage}. The other says that cognition is an
engineered stack---perception here, memory there, an executive on
top~\citep{anderson2004integrated,laird2012soar,lecun2022path}---and that
each function needs its dedicated box. Neither intuition
survives contact with measurement. The purpose of this paper is to put the
boundary itself under the instrument: for each function of a small but
complete cognitive architecture, to determine experimentally whether it
emerges from training or has to be computed by explicit machinery, and to
report the mechanism behind every answer.

\begin{figure}[t]
\centering
\includegraphics[width=0.98\textwidth]{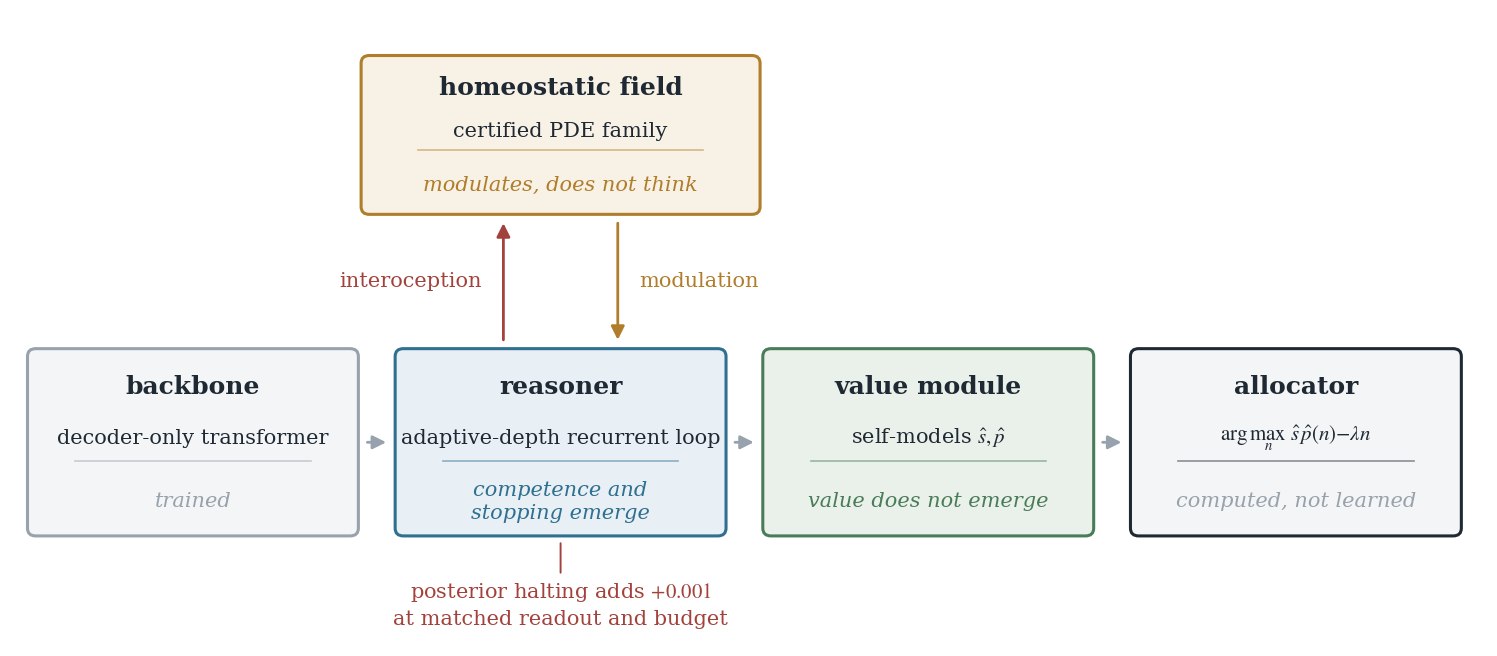}
\caption{The minimal complete architecture, annotated with the program's
verdict for each component. Competence and stopping emerge in the reasoner;
the certified homeostatic field modulates compute without carrying content;
value-shaped decisions do not emerge from any trained coupling and are
computed by an explicit allocator over learned self-models. The posterior
channel---the halting head observing the evolving state---\emph{appears}
worth $+0.223$ over everything decidable ex-ante, but a readout audit
attributes that gap to how adaptive-depth arms are read rather than to what
they know: at matched readout and budget it is worth \Rasign{}
(\S\ref{sec:hierarchy}).}
\label{fig:arch}
\end{figure}

Our testbed is deliberately modest in scale and deliberately complete in
structure (Fig.~\ref{fig:arch}). A decoder-only transformer backbone feeds a
recurrent \emph{reasoner} with adaptive halting in the style of
PonderNet~\citep{banino2021pondernet,graves2016act,dehghani2019universal}; a
low-dimensional homeostatic field---the subject of a companion
paper~\compcite{}---modulates its compute; an explicit slot
memory and a value module complete the system.

 Tasks are algorithmic
($S_5$ permutation composition and relatives), chosen because their difficulty
and information structure can be controlled exactly, and because fixed-depth
transformers provably struggle with them and shortcut solutions fail to
length-generalize~\citep{merrill2023parallelism,liu2023shortcuts,
anil2022length,zhou2024algorithms}. In the final chapters we
swap the trained substrate for a frozen 14B-parameter language
model~\citep{yang2024qwen25} acting as a text actuator, which lets us ask the
same questions of test-time compute in the
LLM regime~\citep{wei2022chain,wang2023selfconsistency,snell2024scaling}.

The paper makes five contributions. \textbf{(1)} An \emph{information
hierarchy} for compute allocation, measured at matched compute on frozen
solvers: uniform \Hun{} $<$ difficulty \Hdif{} $<$ ex-ante value \Hval{},
which is the hierarchy's honest ceiling. The apparent fourth rung, posterior
self-observation at \Hpost{}, is \emph{not} commensurable with the three
below it---it is read from a halting-weighted mixture of states that no
ex-ante arm shares---and at matched readout and matched budget, knowing which
instance needs how much compute is worth \Rasign{} \RasignIC{}
(\S\ref{sec:hierarchy}). \textbf{(2)} The \emph{computed-decision
thesis}: trained couplings leave the entire value-shaped payoff on the table
while an explicit argmax allocator over learned self-models captures all of
what the stake-class oracle can reach---an equality that is partly by
construction, since our stake sensor is perfect (\S\ref{sec:decidir}). \textbf{(3)} The \emph{precondition of governance}
on LLM actuators: before asking how to govern test-time compute one must
measure whether the compute lever has dynamic range at all, and for
self-consistency voting on our actuator it barely does---what any stake-aware
allocation can buy is a two-sided bound, above zero and below our relevance
threshold, with a
mechanism (\S\ref{sec:llm}). \textbf{(4)} A \emph{negative space} of twelve
dead or suspended hypotheses (Table~\ref{tab:negative}), each retired with its
mechanism rather than a bare null
(\S\ref{sec:negative}), produced by a protocol we argue is a contribution in
itself (\S\ref{sec:method}). \textbf{(5)} The \emph{cliff result}: executing
our own preregistered prediction about all-or-nothing cost landscapes, with
an outcome that confirms the headline hypothesis, fails to reproduce the
strong form of the prediction (with an interval too wide to refute it
either), and fails its own second preregistered hypothesis in its
preregistered form (\S\ref{sec:cliff}).

\paragraph{Common experimental setup.} Throughout
\S\ref{sec:pensar}--\S\ref{sec:hierarchy}, instances are length-$K$
composition problems over $S_5$ (compose $K$ group generators and report
the resulting permutation; out-of-distribution means $K$ beyond the
training range), solved by decoder-only transformers of $4$--$6$M
parameters (per-variant counts in the repository). A \emph{solver} is a trained checkpoint evaluated frozen
(forward-only); the twelve solvers of \S\ref{sec:decidir}--%
\S\ref{sec:hierarchy} are six seeds $\times$ two independent runs, our
standing replication unit. Each instance carries a \emph{stake}
$s\in\{1,8\}$ with $P(s{=}8)=0.15$, assigned independently of content and
visible only as metadata; \emph{payoff} is stake-weighted accuracy,
$\sum_i s_i c_i / \sum_i s_i$, on a common frozen evaluation set of $16$k
instances. An \emph{arm}---in the bandit sense~\citep{gittins1979bandit,
lattimore2020bandit}---is an allocation policy mapping instances to
reasoner iterations under a shared compute budget, compared at matched
mean iterations. A \emph{kill-gate} is a cheap preregistered go/no-go
measurement empowered to cancel an expensive experiment; design
calibration was limited to at most three predeclared adjustment rounds
per task family, after which the design froze.

\paragraph{Relation to prior work.} Our questions descend from rational
metareasoning: the hierarchy of \S\ref{sec:hierarchy} is an empirical
measurement, at matched compute, of the value of computation in the sense
of \citet{russell1991metareasoning}---a lineage running through
\citet{horvitz1987reasoning} and \citet{hay2012selecting}---and its
resource-rational reading follows \citet{lieder2020resource}. Adaptive-depth and conditional-compute architectures
\citep{graves2016act,banino2021pondernet,dehghani2019universal,
elbayad2020depth,schwartz2020right,raposo2024mixture} supply our
stopping machinery; we measure \emph{which information} makes stopping and
allocation pay, not only how to learn them. The test-time-compute
literature---voting, verifiers, search, and repeated
sampling~\citep{wei2022chain,wang2023selfconsistency,cobbe2021training,
lightman2024let,yao2023tree,madaan2023self,brown2024monkeys,
snell2024scaling}---supplies the levers of
\S\ref{sec:llm}--\S\ref{sec:cliff}; our
contribution there is the precondition---measuring a lever's dynamic
range, and the reliability of self-agreement
\citep{kadavath2022know,lin2022teaching}, before optimizing over it. The mechanistic facts we lean on, induction
heads and grokking, are from
\citet{elhage2021mathematical,olsson2022induction,power2022grokking,
nanda2023progress}.

\section{The substrate and its certification}\label{sec:pensar}

Before asking where cognition lives one must be sure there is cognition in
the building. We resist defining ``reasoning'' by decree. In this program a
\emph{reasoner} is not a module but a \emph{regime} of a
(task, system, training) triple, certified by a battery of four measurements
we call T-M-I-P: the \textbf{T}ask demands serial computation; the
\textbf{M}emory of the trajectory carries content; halting is
\textbf{I}nformed; and de\textbf{P}th extrapolates. The load-bearing
test is M. If the recurrent state actually carries the computation, then
transplanting the state mid-trajectory between two instances must transplant
the answer. It does: in state-swap interventions the prediction follows the
donor state in $0.83$--$0.92$ of cases (the \emph{cycle} family, which cleared
the seriality gate in two of three seeds; the \emph{adjacent} family failed it
outright here and carries no cognitive claim in this paper, although it
remains a valid compute-allocation benchmark and is, as it happens, the family
in which the order effect of \S\ref{sec:gobernar} survives---a tension we flag
rather than resolve) and never spontaneously
reverts to the recipient's answer ($0.00$),
and representations crystallize over ticks (readout fidelity $0.5 \to
0.85$--$0.89$ across seeds).
Informed stopping exists as well ($\mathrm{AUC}\ 0.79$--$0.90$), though
certification was not linear: the battery initially \emph{failed} in all
twelve of its preregistered cells against
its preregistered thresholds, an earlier inverted-AUC measurement proved
to be a ceiling artifact, and the passing numbers come from preassigned
remedies and a re-certification---an ordering we report as part of the
audit trail.

The battery also earned its keep negatively. An apparent side effect of
integrating the control field---a small apparent accuracy cost (``the
touch'') of $-0.055$, $p=0.011$---was not confirmed by a preregistered
replication with fresh contemporary controls: the replicated effect was
\FtresARtoque{} \FtresARIC{}---an interval compatible both with zero and
with the original effect---so the finding rests on evidence no stronger
than single-run-per-cell noise ($\sigma_{\mathrm{run}}\approx
0.04$--$0.05$), and we retired it. Out of that episode came a standing
rule of the program: effects below ${\sim}0.05$ require at least two
independent runs per cell, because seed-pairing does not pair run noise. We
will meet this rule again in \S\ref{sec:method}.

\section{The homeostatic governor: substance no, structure only in part,
certifiability yes}
\label{sec:gobernar}

The architecture's most speculative component is a low-dimensional
homeostatic field defined on the graph of the model's modules and driven by a
family of partial differential equations on the graph Laplacian---damped wave
(forced Klein--Gordon), its diffusive limit, and a KdV-type
dispersion---inspired by the brain's neuromodulatory systems
\citep{barrett2017allostasis,man2022homeostatic,vecoven2020neuromodulation}.
The full formalism, its stability certificates (an antisymmetric-placement
lemma with a flutter threshold $\beta\rho(A^3)<2\zeta\omega_0^2$ that is exact
when the stiffness is a multiple of the identity and a design guide otherwise,
discrete Schur--Cohn certificates, an unconditionally stable IMEX
integrator), and the complete experimental record are in the companion
paper~\compcite{}. Here we need its conclusion, because the
rest of the program builds on it: \emph{substance no, structure only in
part, certifiability yes---the field modulates, it does not think, and a
learned governor of matched interface modulates just as well.}

Substance, no: the \emph{type} of physics is irrelevant to accuracy. Wave,
diffusion, gated mixtures, nonlocal Poisson coupling, and even a 2D
Navier--Stokes flow substrate all reach the same accuracy, for a measured
reason: the training gradient coarse-grains time, laminating every demand for
modulation into a quasi-static set-point; and an incompressible flow cannot
concentrate information at a point ($\nabla\!\cdot u=0$ conserves area).
Making the field the \emph{source} of computation is catastrophic
($d_z=$~\FdosDz{} worse than a GRU of matched interface). Structure, in part:
in one of the two generator families---and, once capacity is equalized, only
there---the \emph{second-order} character of the dynamics---inertia---confers
on out-of-distribution compute \emph{allocation} a robustness that the
non-stateful learned halting control lacks (the contrast against the
field-free control below does not by itself isolate order from the presence of
a field): an initial exploratory signal (correlation \VtresCorr{}
versus \VtresCorrBase{} on the original task) that replicated under the
companion's preregistered v3 protocol at smaller magnitude ($0.143$ vs.\
$0.002$ and $0.268$ vs.\ $0.213$ across the two generator sets; paired
$p{=}0.025/0.028$, directionally consistent in both but not surviving Holm-4
individually) and survived an A/B against a numerics bug that had frozen the
physical constants. A reviewer-requested deconfounding campaign (twenty
fresh seeds; first order with equalized caps; a matched-interface GRU
replacing the integrator) then delivered the chapter's final, humbler
form: the order effect is strong in the \texttt{adjacent} generator family
($+0.087$, $[+0.042,+0.132]$, $t{=}4.0$, $p{=}3.5{\times}10^{-4}$) and is not
detected in \texttt{cycle\_transp} once capacity is equalized ($+0.014$,
$[-0.013,+0.040]$, n.s.), so under Holm-2 the order hypothesis is not
confirmed as a family-level claim; and the learned GRU governor of matched
interface ties the field in \texttt{adjacent} ($+0.006$, $[-0.051,+0.062]$,
n.s.) and is nominally \emph{better} in \texttt{cycle\_transp} ($-0.035$,
$[-0.067,-0.002]$), with mutual non-inferiority of long-stratum accuracy at a
$0.02$ margin. What survives is the
value of a stateful modulating governor with this interoceptive
interface---of which the field is the \emph{certifiable}
implementation---rather than any superiority of the physics.

The field's final audit---the evidence-accumulator kill-gate---is reported
in full in the companion; here we need its verdict. The last live
hypothesis for making the field load-bearing was that it could serve as a
temporal \emph{evidence accumulator}---a damped second-order filter integrating the
reasoner's noisy per-tick posterior stream, the physical form of sequential
analysis. A kill-gate whose primary cell and $0.03$ pass threshold were fixed
in a dated design document before the verdict (not a formal preregistration;
the probe was recalibrated after a first null) tested the premise on twelve frozen
solvers with a certified probe (its positive control detected planted
distributed signals at the threshold scale): does the posterior stream carry
information about success beyond the last tick? It does not:
$\Delta\mathrm{AUC}=$~\NGunoDelta{} \NGunoIC{} in the predeclared primary
cell, with every secondary cell within $[-0.0012,+0.0073]$---none near the
$0.03$ threshold. The mechanism is structural. The recurrent state already
integrates its own history; the last tick is a sufficient statistic, and the
accumulator's niche is occupied by construction. With that, every employment
imagined for the field is measured, and none is performed better by the
field than by a dedicated component---not even compute control, where a
matched-interface GRU ties it in one generator family and is nominally better
in the other. What is left to the field is not a performance niche but a
licensing one: it is the only implementation of this governor that ships with
a placement dichotomy, a flutter threshold and an unconditional IMEX bound.

\section{Second-order decisions do not emerge; they are computed}
\label{sec:decidir}

The center of the program is a series of integration experiments asking where
decisions about \emph{what matters} and \emph{how much to spend} come from.
(We call these \emph{second-order decisions}---decisions about the
reasoning process, not within it; no relation to the second-order
\emph{dynamics} of \S\ref{sec:gobernar}.)
The design rule throughout---learned the hard way---is to force capability
through the \emph{environment}, never through the loss.

\paragraph{Anticipation has no exploitable value in a smooth family.} We
built session environments---sequences of instances whose hidden difficulty
regime switches stochastically, under a hard shared compute budget---where
anticipating the regime from history could in principle pay, and scanned
$31$ configurations of stakes, budgets, horizons, and regime dynamics
against an oracle ladder (up to a perfect Bayes filter over the whole
history). The ceiling for anticipation was $\le +0.001$ payoff: the value
profile is smooth and regimes mix faster than any anticipatory policy can
exploit, so the reactive stationary policy is already within $+0.001$ of the
perfect-Bayes ceiling in every configuration we scanned. No GPU was
spent on a confirmatory experiment that could not pay.

\paragraph{Gradient coupling is deaf, and weighted losses self-defeat.} We
then handed the trained system every advantage: a value channel with a
near-perfect sensor (a learned stake predictor reaching $\mathrm{AUC}=0.95$;
$1.00$ from raw embeddings) wired into the halting mechanism, and trained
end-to-end with value at stake. The behavioral result: routing correlation
$-0.01$ in every arm---the coupling captures nothing. The mechanism is
legible, and we suspect general though we measured it only here, and worth
stating: \emph{training suppresses task-irrelevant information from the
state} (a probe recovers stake from raw embeddings at $1.00$ but from the
trained state at only $0.79$); the state serves the loss, and anything the
loss does not need decays. The complementary failure is starker. Weighting
the cross-entropy by stakes---making the loss ``feel'' consequences---makes
the model \emph{worse on the very instances it up-weights} ($0.786$ versus
$0.931$ for a stake-blind control; effective sample size \NdosESS), and the
blind control is better by \NdosBlind{} overall. Consequences in the loss are
not a teacher; they are a variance machine. (Both dissections are exploratory branches that the preregistration did
not anticipate, and we report them as such; the preregistered primary
contrast---value-coupled versus uncoupled halting---was an exact null,
$\Delta = +0.0000$.)

\paragraph{The explicit allocator captures the entire ex-ante ceiling.} The
same payoff that gradient coupling captured at zero is then captured
completely by a computed decision: $n^* = \arg\max_n \hat{s}\cdot
\hat{p}(\mathrm{success}\,|\,n) - \lambda n$, where $\hat{s}$ and $\hat{p}$
are self-models fitted from the system's own probes, and the argmax runs at
evaluation time on frozen solvers. Against a difficulty-only baseline the
allocator gains \AunoDelta{} ($t{=}24.97$ against the preregistered relevance
threshold $\delta_0{=}0.02$, not against zero, paired at the seed level
($n{=}6$, $\mathrm{df}{=}5$), one-sided $p=8.7\times10^{-7}$; all twelve
replicas positive, hence all six seed-level means positive) with routing
correlation \RoutingExpl{}, and it reaches the
class-information oracle---an allocator given the true stake class---to
three decimals: one hundred percent of the ceiling attainable from ex-ante
information. That equality is partly by construction (the stake sensor is
perfect here, so $\hat s$ coincides with the stake); the non-trivial
learned content is the success model $\hat p$, whose removal---a hand rule
without the learned self-model---leaves $+0.06$ on the table. This is the
computed-decision thesis: in this setting---where value is orthogonal to
content by construction---second-order decisions do not emerge from the
gradient; they are computed, and computing them works at the limit of the
information our probes expose. Whether the verdict survives when value is
correlated with content is untested.

\section{The information hierarchy}\label{sec:hierarchy}

\begin{figure}[t]
\centering
\includegraphics[width=0.92\textwidth]{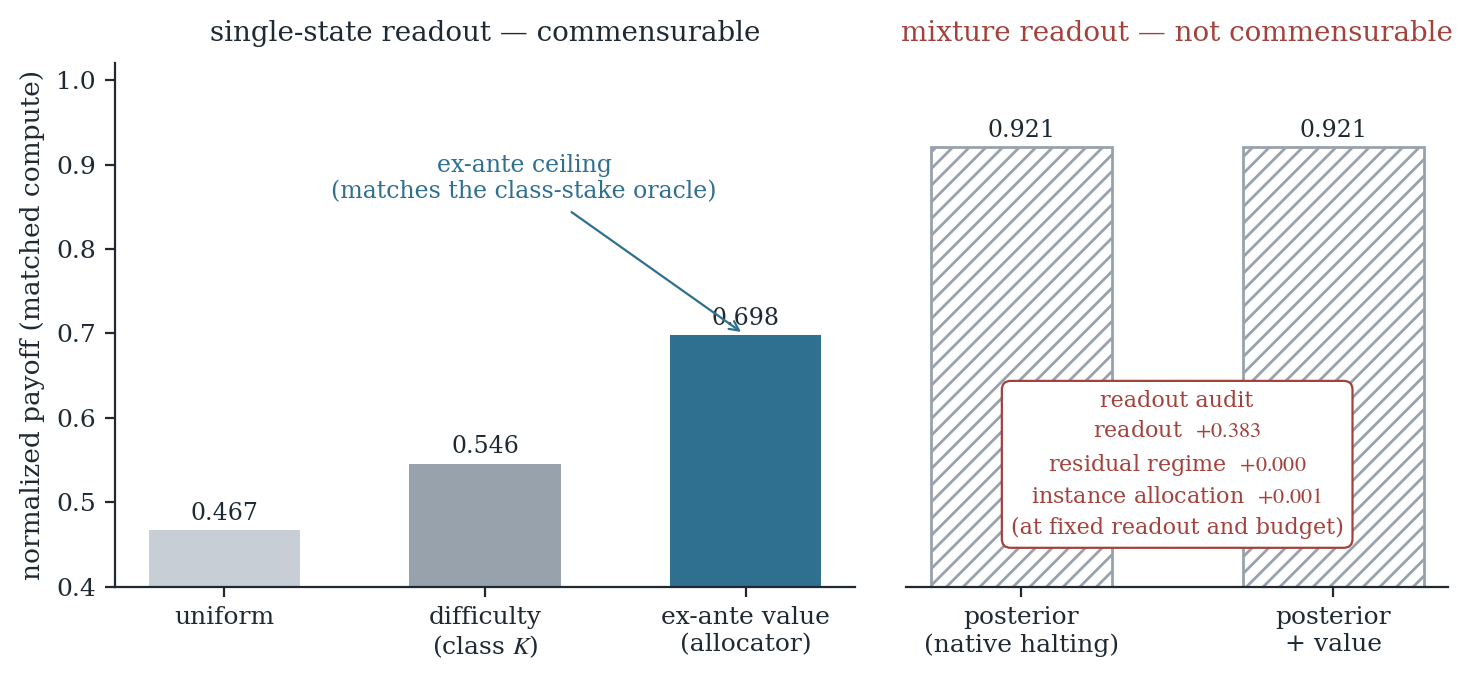}
\caption{The information hierarchy on the trained substrate: normalized
payoff at closely matched mean compute (ex-ante arms at $e{=}5$ exact,
native arm at its emergent $\bar n{=}5.47$, i.e.\ $9\%$ more compute; see the
execution-regime caveat
in \S\ref{sec:hierarchy}), on twelve frozen solvers with a common 16k
evaluation. The explicit ex-ante allocator
reaches the class-stake oracle to three decimals, partly by construction. The top rung is \emph{not}
commensurable with the other three: it is read from a halting-weighted
mixture of states while the ex-ante arms are read from a single state, and
the readout audit of \S\ref{sec:hierarchy} attributes the whole apparent
$+0.223$ to that difference (residual execution regime \Rregimen{}; instance
allocation at fixed readout \Rasign). Value on top of the posterior buys
\VGBcero.}
\label{fig:hierarchy}
\end{figure}

Putting the arms side by side at closely matched compute (exact at $e{=}5$
among the ex-ante arms; the native arm runs at its emergent $\bar n{=}5.47$)
yields the program's central
figure (Fig.~\ref{fig:hierarchy}). Uniform allocation earns \Hun. Knowing
task difficulty earns \Hdif. Knowing ex-ante value---through the explicit
allocator of \S\ref{sec:decidir}---earns \Hval, which equals the class-oracle
ceiling. Letting the system \emph{observe itself think}---the native
adaptive halting reading the evolving state and stopping on its own learned
schedule---\emph{appears} to earn \Hpost, read, crucially, through a readout
the other three arms do not share (next paragraph). And biasing that native
stopping with value, at matched
expected compute and with a preregistered sweep of bias strengths, adds
\VGBcero{} (mean $\pm$ standard \emph{deviation} across the twelve solvers;
the standard error is $0.0001$):
nothing.

The $+0.223$ does not survive audit, and the audit is the chapter's
sharpest result. A \emph{yoked} control, run at reviewer request, forced
execution at exactly the per-instance depths the native halting chose
(rounded $\mathbb{E}[n_i]$) and recovered almost none of the native
advantage: against uniform forced allocation the native arm gains \Ytotal{}
\YtotalIC{}, of which transplanted depths accounted for only \Yinfo{}
\YinfoIC. We first read the remaining \Yregimen{} as the value of elastic,
on-policy execution. It is not. It is the readout.

\paragraph{The readout audit.} \texttt{AdaptiveHalting} returns
$\sum_n p_n x_n$, a halting-weighted \emph{mixture} of hidden states;
the forced path returns the single state reached after the quota; and the
language head was trained with cross-entropy on the mixture, so it has
never decoded a single state. Reconstructing every arm post hoc from the
\emph{same} recorded native trajectory separates the two (twelve solvers,
common evaluation; the reconstruction reproduces the native arm exactly,
and the post-hoc single state at $\hat n$ agrees with the independently
re-executed forced arm on all twelve). The decomposition is exhaustive to
machine precision: the readout accounts for \Rlectura{} \RlecturaIC, the
residual execution regime for \Rregimen{} \RregimenIC, and transplanted
depth information for \Yinfo{} \YinfoIC. Forcing execution costs nothing:
it reproduces the native trajectory tick by tick. Reading it differently
costs everything. The same artifact explains the predeclared VG-N3d
control, which had measured $-0.34$ for forced-depth execution and
attributed it to the cost of commitment.

\paragraph{With the readout held fixed.} The audit invalidates the
native-versus-forced comparison but not the question behind it, which we
then ask cleanly: keep every arm inside the mixture family, over the same
recorded states, and vary only the weights. Giving each instance the
halting distribution of a \emph{different} instance---a derangement, which
preserves the mean budget by construction to four decimals---costs
\Rasign{} \RasignIC{} (a second, independent derangement: $+0.0006$,
$[-0.0004,+0.0016]$, straddling zero). The permutation is not inert: it displaces each
instance's budget by \Rdesplaz{} ticks on average, against a
between-instance standard deviation of $1.26$. What the mixture \emph{is}
worth, relative to a flat uniform mixture at the same mean budget, is
\Rforma{} \RformaIC---the learned population-level shape, not knowledge of
the instance in front of it. So the posterior neither knows nor acts: at
matched readout and matched budget, knowing which instance needs how much
compute is worth one part in a thousand.

Two readings follow. First, the hierarchy's honest ceiling is the ex-ante
rung: what the stake-class oracle can decide in advance is captured in full by
the explicit allocator, and in this architecture self-observation adds
essentially nothing beyond an artifact of how \emph{our} adaptive-depth arms
are read---a claim about PonderNet-style mixture readouts, untested in other
adaptive-depth designs. Second, this
retroactively explains \emph{why} value on top of the posterior buys
\VGBcero{}: we had read that null as elastic stopping already giving each
instance what it needs, but the truth is plainer---per-instance allocation
barely matters here at all, so there is nothing for value to improve.
Value earns its keep only where allocation must be
decided before observation is possible---a scope that the cliff experiment
of \S\ref{sec:cliff} makes precise.

\section{The precondition of governance: a frozen LLM actuator}
\label{sec:llm}

Chapters above measure a substrate we trained. Modern practice, however,
governs \emph{frozen} language models at test time, most commonly by sampling
$n$ chains of thought and majority-voting
\citep{wei2022chain,wang2023selfconsistency,brown2024monkeys}. We connected our instruments to
a frozen Qwen2.5-14B-Instruct actuator~\citep{yang2024qwen25}: tasks verbalized
into text, stakes carried in metadata the actuator never sees, and every
allocation policy evaluated offline over a single cached sample pool, so all
arms are deterministic functions of the same generations.

The result reframes the question. Before asking \emph{how} to govern
test-time compute one must ask whether the lever has dynamic range, and here
it has very little: real, measurably above zero, and bounded well below
relevance. Across seventeen cells spanning two task families and base
accuracies from $0.13$ to $0.96$, the probability that the modal answer of
sixteen samples is wrong obeys an empirical law
$P(\mathrm{modal\ error}) \approx (1-\mathrm{acc}_1) - \varepsilon$ with
$\varepsilon \le$~\LLMepsilon{} in sixteen of the seventeen (the exception,
the highest-slope cell, reaches $0.083$): \emph{the modal answer is
essentially the answer of a single sample}, because errors are systematic---the model agrees
with itself while being wrong (modal-error mass \LLMmodalerr{} in the frozen
cell). Temperature partially decorrelates the errors, multiplying the vote-curve
slope sevenfold and enlarging what lever exists---the
\emph{value-allocation ceiling}, the maximum payoff any stake-aware
allocation of samples can gain over uniform at matched average samples,
computed exactly from the measured vote curve with an unbiased
without-replacement estimator, rises from about \LLMtechoTbaja{} at $T{=}0.7$
(a dial-round figure recorded only in the findings document and not re-derived
with the corrected estimator) to
\LLMtecho{} at $T{=}1.3$---but even then
the ceiling is a two-sided bound: significantly above zero and significantly
below our preregistered relevance threshold of $0.04$
(\LLMtecho{} \LLMtechoIC{}, unbiased vote estimator, paired bootstrap).
With that lever, the governance question does not arise---which is itself
the finding: \emph{the lever is a measurable property of the (actuator,
spending mechanism) pair, and measuring it comes first.}

\paragraph{An anchor on a public benchmark.} A natural objection is
that self-consistency famously gains $10$--$18$ points on
GSM8K~\citep{wang2023selfconsistency}, which would put
$\varepsilon\approx0.1$--$0.3$, far above our law. At reviewer request we
ran the anchor cell: $256$ instances of the official GSM8K test with the
same actuator and instruments ($m{=}16$, $T{=}0.7$). The law holds:
$\mathrm{acc}_1=$~\GsmAccUno, the vote curve peaks at $0.941$ ($n{=}7$--$9$,
slope \GsmPend{} to $n{=}15$), modal-error mass is \GsmModalErr---about
$70\%$ of the remaining error is systematic---and
$\varepsilon=$~\GsmEps~$\le0.08$. There is no contradiction with the
published gains: those were measured on much weaker samplers
($\mathrm{acc}_1\approx0.4$--$0.6$) whose error mass was large and
diversifiable, while a strong modern instruct model leaves little error
and most of it systematic. The precondition finding---measure the
lever's range before optimizing over it---carries to the public
benchmark unchanged.

The chapter's sharpest result is a contrast of posteriors. On the trained
substrate, self-observation \emph{appeared} to be the greatest asset
($+0.223$) until the readout audit reduced it to \Rasign{} at matched
readout and budget. On the sampled LLM the analogous channel, inter-sample
agreement, is nearly worthless too---and the two now agree. With paired bootstrap intervals computed
at reviewer request, stopping on agreement buys $+0.017$ at most in point
estimate (upper confidence limit $+0.025$) over
ex-ante allocation ($+0.007$ $[-0.008,+0.015]$, $+0.015$
$[+0.001,+0.021]$, $+0.017$ $[+0.005,+0.025]$ at matched budgets of
$\bar n \approx 3, 5, 7$ samples; an earlier point estimate had the sign
negative, and the interval analysis corrects both the sign and the
emphasis)---as close to worthless as the audited posterior advantage on the
trained substrate, though an order of magnitude larger than it in absolute
terms ($+0.017$ versus \Rasign{})---while agreement
mass concentrates on wrong
answers (modal-error \LLMmodalerr): agreement measures conviction rather
than correctness, a failure mode distinct both from eliciting calibrated
self-knowledge, which these models largely
have~\citep{kadavath2022know,lin2022teaching}, and from external
verification~\citep{cobbe2021training,lightman2024let}. Value on top of
the LLM posterior is likewise null with intervals
($+0.003$/$+0.002$/$+0.001$, all straddling zero). The same
act---watching oneself think---is nearly worthless in \emph{both} regimes.
That convergence was invisible while one of the two was measured through a
readout the other did not share, which is the methodological point of this
chapter.

\section{Negative space as a map}\label{sec:negative}

A null without a mechanism is noise; with one, it is cartography. The
program's negative results, each with the measurement that explains it, are
collected in Table~\ref{tab:negative}. Together they draw the boundary that
the positive results inhabit: the field does not think, the gradient does
not route, the loss does not teach values, self-observation does not
allocate, the sampled LLM's agreement does
not verify---and the places where each function actually lives are exactly
the ones described in \S\ref{sec:gobernar}--\S\ref{sec:llm}.

\begin{table}[t]\centering\small
\begin{tabular}{@{}>{\raggedright\arraybackslash}p{0.29\textwidth}
>{\raggedright\arraybackslash}p{0.25\textwidth}
>{\raggedright\arraybackslash}p{0.38\textwidth}@{}}
\toprule
Dead hypothesis & Evidence & Measured mechanism\\
\midrule
Field as compute source & $d_z$ = \FdosDz{} vs GRU & plan readout fails to extrapolate; inertia lags budget shifts\\
Field physics as load-bearing & $|\Delta|\le0.012$ ($\times$3 lines) & gradient coarse-grains time into set-points\\
Multiscale anticipation ($S_5$) & $\le+0.001$ (31 configs) & smooth value; regimes mix faster than foresight pays\\
Routing via gradient & corr $-0.01$ (AUC $0.95$) & state sheds task-irrelevant information\\
Consequences in the loss$^\dagger$ & blind \NdosBlind{} better & ESS \NdosESS: weighting destroys the solver\\
Value atop the posterior & \VGBcero & per-instance allocation barely matters\\
Posterior allocation as information & \Rasign{} \RasignIC{} (readout fixed) & the mixture readout, not knowledge of the instance\\
The integration ``touch'' & \FtresARtoque{}~\FtresARIC & unconfirmed; interval spans zero and the original\\
Field as evidence accumulator & \NGunoDelta{} \NGunoIC & recurrence already integrates\\
Self-consistency as lever & ceiling \LLMtecho{} $<0.04$ & systematic errors; modal $\approx$ 1 sample\\
LLM agreement as stopping & $\le+0.017$ vs ex-ante (CI) & measures conviction, not correctness\\
Cliff family \emph{constructible} on the small substrate (the prediction itself stays open there) & stuck at prior, 3 predeclared rounds & in-context retrieval blocked by the loop\\
\bottomrule
\end{tabular}
\caption{The negative space: where cognition does \emph{not} live. All but one
entry is a measurement with a mechanism rather than a bare null (the
integration ``touch'' is a non-replication, reported as unresolved); rows are
preregistered confirmatory results except the one marked $\dagger$, an
exploratory branch the preregistration did not anticipate, and two audits run
post hoc at reviewer request (posterior allocation; the LLM agreement
intervals), all reported as such.}
\label{tab:negative}
\end{table}

\section{Method as a result}\label{sec:method}

The findings above are only as good as the process that produced them, and
the process had to save us from ourselves repeatedly; we therefore report it
as a contribution, in the spirit of the reproducibility
literature~\citep{nosek2018prereg,ioannidis2005why,simmons2011false}. Every
confirmatory experiment was
preregistered with complete verdict branches before data. Every design was
attacked by an adversarial review panel before GPU was spent---across the
program, seven panels produced more than two hundred findings, including genuine
code bugs found by reviewers reading sources. Cheap kill-gates ran before
expensive confirmatories, and calibration was firewalled at three declared
dial rounds per family. Three episodes illustrate why none of this is
ceremony.

\emph{Replication.} The ``touch'' of \S\ref{sec:pensar} was significant at
$p=0.011$ and was not confirmed on replication---the replication interval
covers both zero and the original effect---which is precisely why the
replicas rule (two runs per cell for small effects) is now standing
policy.

\emph{No null without a positive control.} Twice in two days an instrument
produced a false verdict in the null direction. The vote-curve estimator for
the LLM chapter sampled with replacement from the cached pool,
underestimating exactly where pools are diverse, and its ceiling was
evaluated at a design cap rather than the affordable budget. Correcting both
turned a ceiling once believed structurally unreachable into a real,
measurable---but still sub-threshold---lever; a hasty re-measurement on $96$ instances without a paired
interval then produced a transient false positive ($+0.050$, above
threshold)---our third error---before the full paired probe settled the
honest two-sided bound of \S\ref{sec:llm} ($+0.0236$, significantly above
zero and significantly below $0.04$). Out of the episode came a second
standing rule: no verdict without an interval on the difference. Days
later, the evidence-accumulator probe of
\S\ref{sec:gobernar} initially paid an overfitting tax that biased it
toward null; a planted-signal control caught it before the verdict. The
resulting rule---no null is reported unless its instrument demonstrably
detects a planted effect of the threshold size---did the most work of any
single practice in the program.

\emph{Pre-declared controls block bad instruments.} An offline pilot of the
cliff experiment burned three instrument designs in sequence; each was
blocked by the same preregistered sanity condition (reproduce a known zero)
before any conclusion could be contaminated, and the third failure revealed
a structural fact about the known zero itself that redesigned the real
experiment.

\section{The cliff: executing our own prediction}\label{sec:cliff}

Three independent chapters---anticipation, routing, and value atop the
posterior---died against the same two properties of the smooth family:
smooth value and an allocation problem in which which-instance-gets-what
barely matters (\S\ref{sec:hierarchy}). That convergence licensed a
falsifiable prediction: in a \emph{cliff} family, where task cost is
all-or-nothing and invisible from inside, the hierarchy should tilt back
toward ex-ante decisions. This section reports the execution of that
prediction, in three acts.

\paragraph{Act one: the family resists construction on the small substrate.}
Our cliff family is \emph{cycle walking}: given a random permutation
presented as shuffled pairs, walk from $s$ to $t$ and report the distance
$d$---unknowable until arrival, with only the cycle length $L$ visible
ex-ante. On the trained substrate the family failed its learnability gate
three times, with a three-part mechanism. The task requires \emph{in-context}
retrieval (with the table in the weights, a spectral shortcut computes $d$
in one pass and the cliff evaporates---the Fourier phenomenology of
grokking~\citep{power2022grokking,nanda2023progress}); the induction circuit
that performs in-context retrieval~\citep{olsson2022induction} forms only
under a specific signal regime (dense supervision, large batch: loss sits
at the exact support entropy $\ln 20$ for nine thousand steps, then
transitions to perfect retrieval by fifteen thousand---on the plain
backbone); and the recurrent loop that the family needs for its compute
cliff \emph{blocks that formation} (the same data regime that groks on the
backbone stays at prior through thirty thousand steps under the reasoner).
The component the family needs for the compute prevents learning the one it
needs for the task. (The attribution is convergent rather than surgical:
the round-3 run differs from the plain-backbone probe in one further
respect---the distance-question head---though its auxiliary losses alone
also stayed at $\ln 20$.) Signal regimes, like levers, are preconditions.

\paragraph{Act two: the family's natural home.} A pretrained LLM has
induction heads for free, and its compute lever for this task is generation
\emph{length}, not votes: walking $d$ steps costs tokens proportional to
$d$ (measured correlation $1.00$), the visible class (the cycle length $L$, the only cost information
available before walking) predicts cost poorly in the cliff family
($R^2 =$~\NbLRdosCliff) and well in a smooth arithmetic family
($R^2 =$~\NbLRdosSuave)---the operational definition of a cliff. The
actuator genuinely walks: one hundred percent of correct answers contain a
complete valid hop chain, so cost-proportionality holds literally. Every
allocation arm is an offline truncation of a single recorded greedy
generation per instance, re-parsed token-prefix by token-prefix, with
budgets accounted as tokens \emph{spent} and allocations solved exactly by
dynamic programming.

\begin{figure}[t]
\centering
\includegraphics[width=0.98\textwidth]{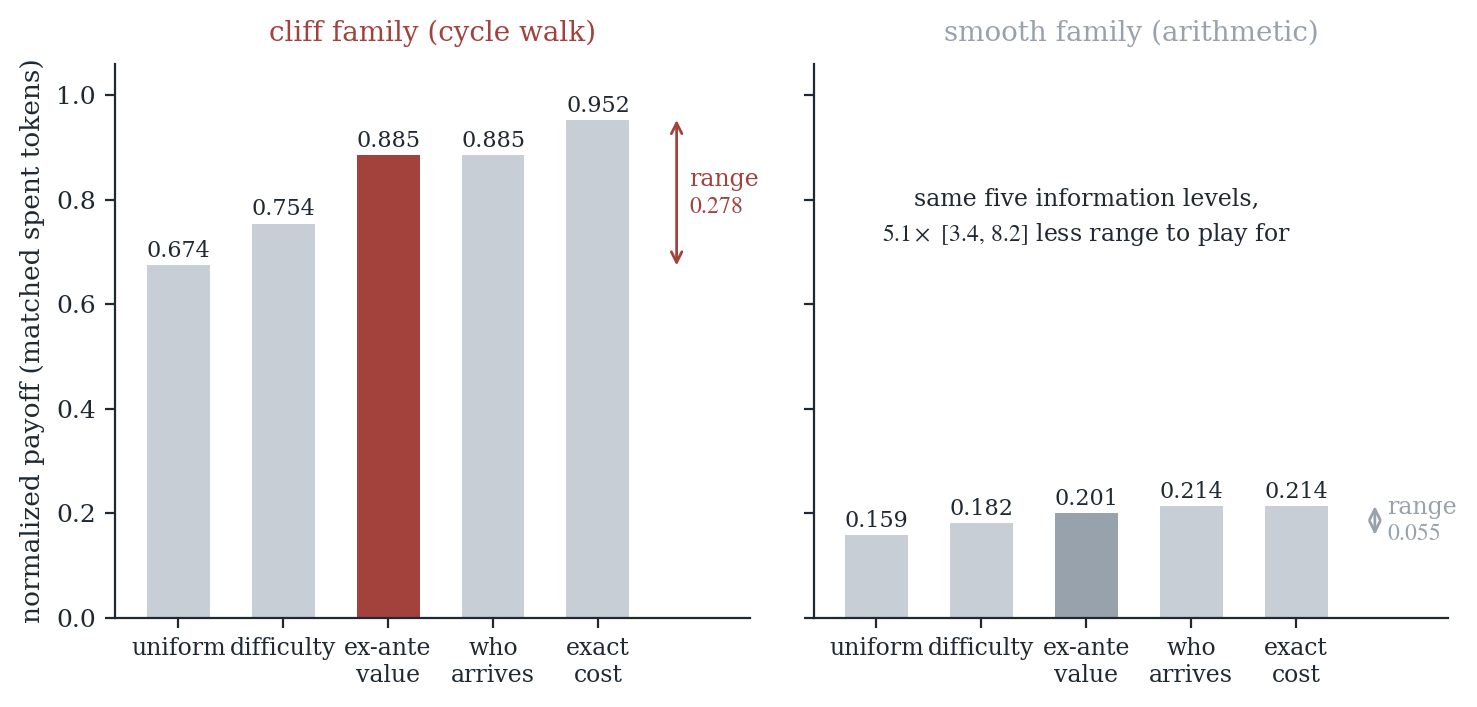}
\caption{The cliff ladders on the frozen LLM actuator (left: cycle walking;
right: arithmetic with visible work). All arms are offline truncations of
the same recorded generations at matched spent tokens. The cliff multiplies
the attainable allocation range fivefold (\NbLRangoCliff{} vs
\NbLRangoSuave) with no \emph{detected} change in the fraction of it that
ex-ante information captures (\NbLFracCliff{} vs \NbLFracSuave{}), a
comparison too imprecise to establish invariance. Bars are point
estimates from single recorded greedy generations; the fraction
comparison is post-hoc with a wide interval (difference \FracDif{}
\FracDifIC), while the range ratio \RatioRangos{} \RatioRangosIC{}
excludes unity decisively (paired instance bootstrap, frozen
allocations).}
\label{fig:cliff}
\end{figure}

\paragraph{Act three: the result.} Figure~\ref{fig:cliff} shows both
ladders. In the cliff family, payoff climbs from $0.674$ (uniform) through
$0.754$ (difficulty) to $0.885$ (ex-ante value), and knowing \emph{who will
arrive} adds exactly nothing ($0.885$), while
knowing the \emph{exact cost} reaches $0.952$. The preregistered headline
confirms with room to spare: value minus difficulty is \NbLHuno{}
\NbLHunoIC{}, $3.4\times$ its threshold, against \NbLHunoSuave{} in the
smooth family---value under commitment pays, and pays most where cost is
invisible in advance. The preregistered opposite-direction contrast also
confirms: the margin left above difficulty-only allocation is
\NbLContrCliff{} in the cliff versus \NbLContrSuave{} in the smooth family.

And then the finding we did not seek. Normalizing each family by its own
attainable range, the fraction captured by the best ex-ante policy is
\NbLFracCliff{} $[0.703, 0.825]$ in the cliff and \NbLFracSuave{}
$[0.522, 0.899]$ in the smooth family; the difference is \FracDif{}
\FracDifIC{} (paired instance bootstrap, frozen allocations)---a post-hoc
observation compatible with equality, though the smooth family's fraction
is imprecisely estimated, so we claim no detected difference rather than
invariance. The range ratio, by contrast, is solid: \RatioRangos{}
\RatioRangosIC{}. The strong form of our prediction said the
cliff would shift the information balance toward ex-ante decisions. It does
not. \emph{We detect no change in how the information is divided---though
that comparison is imprecise---while the cliff demonstrably multiplies,
fivefold, how much there is to divide} (\NbLRangoCliff{} versus
\NbLRangoSuave). The prediction conflated the magnitude of an allocation
problem with the structure of its information; the data settle the magnitude
axis and leave the structural one undecided.

Honesty requires two demotions. Our second hypothesis---that nothing
realizable pays beyond arrival detection---\emph{fails in its preregistered
form}: exact-cost oracle minus arrival oracle is $+0.067$ against a
$0.0056$ threshold. We read that verdict as uninformative rather than
damning---the preregistered comparator is clairvoyant rather than
mid-flight, a flaw the design panel had flagged and we left standing, so
the margin prices a clairvoyance no realizable observer has---but the
formal verdict is a failure and we report it as one. What \emph{is}
realizable is measured and clean: the arrival oracle (a binary oracle of
who finishes within budget) adds exactly nothing, and a redesigned
mid-flight comparator is specified for future work. And strict muteness remains
unadjudicated: the preregistered probe passed but its own positive control
failed in its preregistered form, and under our no-null-without-a-control
rule we do not claim it.

\section{Discussion}\label{sec:discussion}

\paragraph{The architecture that remains.} Assembling the surviving
positives yields a specific design. Competence is trained. Stopping is
native and posterior---the halting head reading the evolving state---but the
readout audit strips it of the role we had given it: its per-instance
choices carry \Yinfo{} when transplanted, less than difficulty information
alone, and \Rasign{} once the readout is equalized and the budget matched.
The honest ceiling of the hierarchy is therefore the ex-ante rung, which the
explicit allocator reaches to three decimals---an equality that is partly by
construction, since our stake sensor is perfect. Value is computed by an explicit allocator
over learned self-models---anticipation, in these families, has nothing to
compute---and its domain is \emph{commitment}: decisions
that must be made before observation is possible, a domain the cliff result
shows the cost structure of the environment can enlarge severalfold (range
ratio \RatioRangos{} \RatioRangosIC{} between our two families). The homeostatic field survives as what the companion paper
called it: a certified compute governor---brainstem, not
cortex~\compcite{}---and it survives as the \emph{certifiable}
implementation of that role rather than the best-performing one, since a GRU
of matched interface governs at least as well.

\paragraph{One law, met three times.} The program kept colliding with the
same shape of fact at different levels. Before asking how to govern, measure
whether the lever has range (\S\ref{sec:llm}). Before asking whether a
family tests a hypothesis, measure whether its signal regime lets the
substrate learn it at all (\S\ref{sec:cliff}, act one). And when a
capability is worth paying for, the payment scales with the ex-ante
\emph{invisibility} of cost, and not---so far as we could resolve it
(\FracDif{} \FracDifIC)---with any change in who holds the
information (\S\ref{sec:cliff}, act three; a reading based on one pair of
families and one actuator, offered as the sharpest available summary
rather than an established law). Preconditions first; the
interesting question is often one level below the one you asked.

\paragraph{Limitations.} The trained substrate is small ($4$--$6$M parameters;
$4.2$--$5.6$M in the measured variants)
and algorithmic by design; the hierarchy's absolute numbers are properties
of our task families, though the ordering of the three ex-ante rungs held in
all twelve solvers (six seeds $\times$ two runs).
The LLM chapters use a single frozen actuator; the governance chapter
evaluates all arms offline over cached sample pools ($768$ instances,
$m{=}16$--$24$ samples per instance depending on the cell), while the cliff chapter uses greedy
decoding with one recorded generation per instance. Replication across
actuators (our preregistered universality hypothesis) was never run,
though the GSM8K anchor cell extends the lever measurement to a public
benchmark on the same actuator. Twelve replicas arise as six seeds
$\times$ two runs, which is anti-conservative if runs share seed noise; we
report per-seed sensitivity where it matters. Strict muteness and the
realizable mid-flight ceiling remain open, with the redesigned comparator
specified. Our stake is orthogonal to content by construction, and the
measured mechanism of the computed-decision thesis is that training sheds
task-irrelevant information from the state; with value correlated with content
the verdict could differ, and we did not test it. The readout audit likewise
concerns PonderNet-style halting whose language head is trained on the
mixture, not every adaptive-depth design. And all governance results concern
compute allocation; nothing here speaks to gradient-time governance.

\section{Conclusion}

We asked, of every part of a minimal complete cognitive architecture,
whether its function emerges or must be computed, and obtained an answer
with an unusual property: it is the same answer at every scale we probed.
What can be observed turns out to be barely worth observing---the posterior
halting head neither knows nor acts: forcing execution at the depths it chose
reproduces its trajectory exactly (residual regime \Rregimen{} \RregimenIC),
its per-instance choices are worth \Rasign{} \RasignIC{} once the readout is
equalized and the budget matched, and what looked like its advantage was the
mixture readout (\Rlectura{} \RlecturaIC). What cannot be observed must be
computed---explicitly, at decision time, from learned self-models, because
gradients shed exactly the information those decisions need. And before
either, the preconditions must be measured: the range of the lever, the
regime of the signal, the visibility of the cost. The architecture of
cognition, in this program's experience, is less a stack of faculties than
a discipline of measurement---and the map of where cognition does not live
turned out to be the most reusable thing we built.

\paragraph{Reproducibility.} All preregistrations (with dated amendments),
findings documents, panel adjudications, solver checkpoints, caches, and
analysis code are in the program repository. Numbers tied to a preregistered
verdict enter the text as macros transcribed from the archived results and
checked against them, with declared exceptions hard-coded from the same files
(the H2 margin of \S\ref{sec:cliff}, the cliff ladder rungs, and the
agreement-stopping intervals of \S\ref{sec:llm}); the $T{=}0.7$ ceiling
\LLMtechoTbaja{} and the exploratory \VtresCorr{}/\VtresCorrBase{} pair are
descriptive and carry no archived verdict file. Remaining descriptive figures
are checked against the findings documents.

\appendix
\section{Substrate, tasks, and payoff}\label{app:setup}
Instances are length-$K$ compositions over $S_5$: the input lists $K$
generator tokens (certified family: cycle generators), and the target at a
fixed answer position is the composed permutation, with dense supervision
of the running product during training. Out-of-distribution means $K$
beyond the training range. Models are decoder-only transformers
(4 layers, $d_{\mathrm{model}}{=}256$, 4 heads, RMSNorm, RoPE, SwiGLU;
$4$--$6$M parameters depending on variant) feeding a recurrent reasoner
block applied up to $N_{\max}{=}24$ ticks under PonderNet-style halting,
with an external slot memory. Training uses AdamW (cosine schedule,
gradient clipping), BF16 on GPU with the physical field parameters pinned
to FP32. Stakes are metadata: two slot tokens prepended to each instance,
with stake high ($s{=}8$) iff the tokens are equal ($P{=}0.15$,
independent of content and length by construction); the loss never sees
them in the confirmatory arms. Payoff is stake-weighted accuracy
$\sum_i s_i c_i / \sum_i s_i$ on a common frozen evaluation set of $16$k
instances. The twelve solvers are six seeds $\times$ two independent
training runs of the stake-blind arm, evaluated frozen.

\section{Hierarchy arms and estimators}\label{app:hierarchy}
Ex-ante arms execute via per-instance forced depth with exact matching at
$e{=}5$ mean ticks: uniform ($n{=}5$), difficulty (depth by class $K$
from an isotonic success table $\hat p(\cdot\,|\,K)$ fitted on a probe
split), rule (two levels by predicted stake, no success model), the
explicit allocator ($n^*=\arg\max_n \hat s\,\hat
p(\mathrm{success}\,|\,n)-\lambda n$, with $\hat s$ a stake head read
from slot embeddings and $\lambda$ set by the budget), and the class
oracle (true stake). The native arm is the trained halting run as
trained. VG-B0 sweeps a stake-conditioned additive offset on the halting
logit ($\delta\in\{0.5,1,2,3.5\}$) at matched
$\mathbb{E}[\bar n]\pm0.05$. The yoked control re-executes every instance
at forced depth $\mathrm{round}(\mathbb{E}[n_i])$ taken from the native
rollout of that same instance. Contrasts are paired per instance and
aggregated per solver ($n{=}12$; $t$-intervals), except the allocator
contrast A1 of \S\ref{sec:decidir}, which is aggregated per seed ($n{=}6$,
$\mathrm{df}{=}5$) and tested against the preregistered threshold
$\delta_0{=}0.02$.

\section{LLM governance cells}\label{app:llm}
The actuator is frozen Qwen2.5-14B-Instruct. Tasks are verbalized
($S_5$-in-text: three named operations applied to a five-token row);
stakes remain metadata carried outside the prompt. For each cell we cache
a pool of $m$ i.i.d.\ sampled chains per instance ($768$ instances, except
the GSM8K anchor cell, which uses the $256$ official test items;
$m{=}16$--$24$; $T\in\{0.7,1.0,1.3\}$, top-$p{=}0.9$), and every policy
is a deterministic function of the pool. The vote curve
$\mathrm{acc}(\text{vote-of-}n)$ uses the unbiased without-replacement
estimator (subsets of the pool are i.i.d.\ draws by exchangeability;
ties split uniformly; non-parses do not vote). The value-allocation
ceiling is the maximum stake-aware gain over uniform at matched mean
samples, computed exactly from the measured curve. Agreement-stopping
policies are evaluated by dynamic programming over recorded sample-walk
states $(k,\text{modal count})$, fitted and evaluated on disjoint
halves; the paired bootstrap ($B{=}200$) resamples instances and repeats
the full pipeline, including frontier construction.

\section{Cliff recording, truncation, and arms}\label{app:cliff}
Cycle-walk instances present a random single-cycle permutation over
letters as shuffled pairs (cycle length $L\in\{6,10,14\}$ visible;
distance $d\sim U[1,L{-}1]$ hidden); the arithmetic contrast family uses
chained integer operations with the operation count visible. One greedy
generation per instance is recorded as full token IDs. The arrival index
$c_i$ is the minimal token prefix containing a \emph{completed} answer
line (binary search over decoded prefixes), and
$\mathrm{correct}(\mathrm{cap})$ re-parses the truncated prefix
literally, which resolves multi-answer and split-digit artifacts
(self-tested against literal re-parsing on a cap grid). Budgets are
accounted as tokens \emph{spent}, $\sum_i \min(t_i,\mathrm{cap}_i)$,
matched across arms; allocations are solved exactly by multi-choice
knapsack dynamic programming over a cap grid; payoffs marginalize the
stake analytically; caps are cross-fitted (A$\leftrightarrow$B halves).
Gates run on the recordings before any hypothesis is computed: R
(cost--distance correlation $1.00$; cost predictability from the visible
class $R^2$ \NbLRdosCliff{} vs \NbLRdosSuave), S ($100\%$ of correct
answers contain a complete valid hop chain), and censoring ($0.4\%$, $0.4\%$,
$0.0\%$ at $L{=}6,10,14$). The
fraction and range-ratio intervals use a frozen-allocation instance
bootstrap ($B{=}2000$) with the exact-cost oracle re-solved per
replicate.

\paragraph{Author contributions (CRediT).} \ifanon\else
\textbf{F.M.A.-C.} (corresponding): conceptualization, methodology, software,
formal analysis, investigation, visualization, project administration,
writing---original draft. \ 
\textbf{F.G.M.}: conceptualization, methodology, formal analysis, validation,
supervision, writing---review and editing. \ 
\textbf{A.A.}: software, data curation, resources, investigation, validation,
writing---review and editing. \ 
\textbf{I.F.}: conceptualization, supervision, funding acquisition, resources,
validation, writing---review and editing. \fi

\paragraph{Data and code availability.} The full program---model code,
preregistrations, run artifacts and the scripts that produce every table and
figure in this paper---is covered by the MIT licence.
\ifanon
\else\ifdefined\coderepo It is available at \coderepo.
\else It is available from the corresponding author, and a public repository
with an archival DOI will accompany the published version.\fi\fi
Every number in the text enters as a macro transcribed from an archived
artifact and checked against it, so each one can be traced back to the file it
came from.

\paragraph{Use of generative AI.} The experiments were designed, executed and
adjudicated by the authors. A large language model was used as a coding and
drafting assistant throughout, including for implementation, for adversarial
review of the manuscript's internal consistency, and for literature search;
all claims, numbers and verdicts reported here were verified by the authors
against the archived artifacts.

\bibliographystyle{tmlr}
\bibliography{refs}

\end{document}